# Deep Learning Based Robot for Automatically Picking up Garbage on the Grass

Jinqiang Bai, Shiguo Lian, *Member*, *IEEE*, Zhaoxiang Liu, Kai Wang, Dijun Liu

*Abstract*—This paper presents a novel garbage pickup robot which operates on the grass. The robot is able to detect the garbage accurately and autonomously by using a deep neural network for garbage recognition. In addition, with the ground segmentation using a deep neural network, a novel navigation strategy is proposed to guide the robot to move around. With the garbage recognition and automatic navigation functions, the robot can clean garbage on the ground in places like parks or schools efficiently and autonomously. Experimental results show that the garbage recognition accuracy can reach as high as 95%, and even without path planning, the navigation strategy can reach almost the same cleaning efficiency with traditional methods. Thus, the proposed robot can serve as a good assistance to relieve dustman's physical labor on garbage cleaning tasks.

## I. INTRODUCTION

MANUAL garbage pickup and cleaning is a tedious, boring and repetitive task [1], and autonomous robot can be a potential candidate for this application. The autonomous floor-cleaning [2], aquatic-cleaning [3] [4], wall-cleaning [5] and rubbish collecting [6] robots have been developed for years, while autonomous cleaning robot that can operate on the grass still remains a challenging task due to the lack of garbage recognition ability.

Motivated by the garbage collection robot working on the beach [7], this paper aims to develop a robot that can identify the garbage and pick it up automatically. As the key component of such robots, automatic trash detection algorithm using ultrasonic sensors was proposed in [8], however, it might confuse an obstacle with garbage due to the limited perception ability of ultrasonic sensors. In contrast, web camera can be a preferable sensor for object detection because it provides much more information than ultrasonic sensors. Besides, deep neural networks (e.g. deep convolutional networks) have been applied with great success to the detection, segmentation and recognition of objects in images [9]. Therefore, we employ a convolutional neural network (CNN) for recognizing and locating the garbage in images captured by a web camera equipped on the robot. Moreover, the network can also be used to detect and segment the ground, which is useful for obstacle avoidance.

In order to clean the relatively big garbage (e.g. glass bottles, plastic, and waste paper) on the grass, a manipulator instead of the vacuum or roller brush used on most existing commercial floor-cleaning robots which may destroy the grass or clean the unnecessary trash (e.g. dust, leaves), is adopted as the cleaning device. Specifically, the vision-based grasping controller [10] is used to command the manipulator for garbage pickup.

Besides these key components, the robot presented in this paper also provides basic functions such as path planning, obstacle avoidance, localization, environment perception, etc. All of these modules make up a novel autonomous garbage cleaning (exactly, picking) robot that can be used for picking up relatively big garbage on the grass in a park or school. One challenge of such task is the pre-unknown navigation goal (i.e. the garbage). The conventional map-based visual navigation method [11] will fail in such task because it must know the accurate location of the navigation goal in advance. Recently, the reinforcement learning based visual navigation method [12] can navigate the robot towards a visual target with a minimum number of steps in indoor environment. However, this method requires a training phase in a simulate environment, which is very time-consuming. Besides, this method may fail in an untrained and real environment. Moreover, the requirement of distinguishing between garbage and non-garbage (treat as obstacles) for selecting the navigation goal makes it more challenging. Thus, a novel navigation strategy based on ground segmentation is proposed in this paper to realize the online and autonomous target selection and avoid non-garbage (obstacles) simultaneously. To the best of our knowledge, this is the first prototype (see Fig. 1) that can perform such tasks in such complex scenarios.

The rest of the paper is organized as follows. Section II presents the overview of the proposed robot system, including the hardware configuration and system architecture. The key methodology of the novel garbage-cleaning robot is presented in Section III, including perception, object tracking, navigation, and so on. Section IV shows some experimental results and demonstrates the effectiveness of the proposed robot system. The conclusions are drawn in Section V.

Jinqiang Bai is with School of Electronic Information Engineering, Beihang University, Beijing, 10083, China (e-mail: baijinqiang@buaa.edu.cn).

Shiguo Lian, Zhaoxiang Liu, Kai Wang are all with AI Department, CloudMinds Technologies Inc., Beijing, 100102, China (e-mail: { scott.lian, robin.liu, kai.wang }@cloudminds.com).

Dijun Liu is with National Key Laboratory of Wireless Mobile Communication, China Academy of Telecommunication Technology, Beijing, 10083, China (e-mail: liudijun@datang.com).

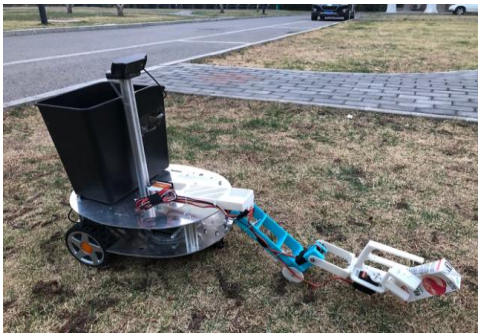

Fig. 1. The prototype of the proposed robot system.

## II. OVERVIEW OF THE PROPOSED ROBOT

This section describes the hardware configuration and system architecture of the proposed robot system.

### A. Hardware Configuration

As shown in Fig. 2, the complete system of this robot includes five major parts: (1) sensors for environment perception, (2) robot base, (3) controller, (4) manipulator, and (5) garbage container.

A low-cost inertial sensor (including a triaxial gyroscope, a triaxial accelerometer, and a triaxial magnetometer), a vehicle odometer, and a GPS (Global Position System) receiver are used to produce continuous and accurate location information. A web camera is adopted for detecting and recognizing the garbage, as well as detecting the obstacle. Since obstacles cannot be accurately detected by the camera alone (such as overexposure or motion blur) [13], an ultrasonic rangefinder is fused to detect obstacle more robustly.

The robot base is the main body of the robot, equipped with all sensors, controller, manipulator and garbage container. The robot base has the dimensions of 400*320*160 mm (length*width*height), and can carry more than 40 Kg. The robot base has two driving wheels and a passive wheel, and is driven differentially to turn left or right.

The controller of the robot is like the human brain, including CPU (Central Processing Unit), GPU (Graphics Processing Unit) and motor driver. All the algorithms such as locating, obstacle avoiding, image pre-processing, and path planning, etc. are performed on the CPU. The garbage recognition and segmentation algorithm are executed on the GPU. The GPU has a parallel structure that makes it more efficient than the general CPU for image processing and deep learning computations. The rotation, forward or back translation of the robot is controlled by the motor driver, which uses the widespread and the mature PID (Proportion Integration Differentiation) algorithm [14].

A manipulator is used to pick up the garbage, and put it back to the garbage container. It consists of four joints with five degrees of freedom. The front-end gripper can grasp objects within a weight of up to 1Kg located within a diameter of 10 cm.

A garbage container is used to collect the garbage, which is fixed on the robot base by a clamping slot type to assemble and dismantle very easily.

### B. System Architecture

The system architecture of the proposed robot includes a localization module, a navigation module, a perception module, a driver module, and a map module, as shown in Fig. 3.

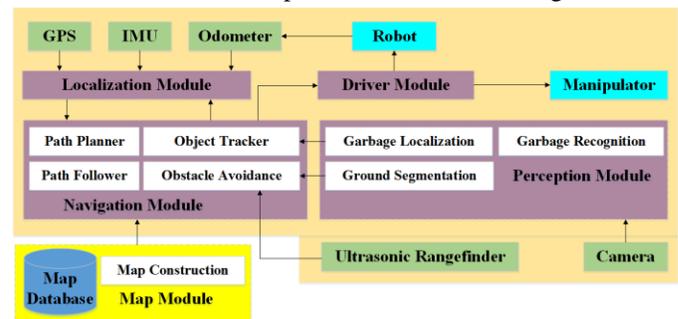

Fig. 3. The system architecture of the proposed robot.

#### 1) Localization module

The IMU (Inertial Measurement Unit) and odometer are integrated into a GPS based autonomous navigation system [15] to locate the robot robustly. The IMU and odometer give the system a dead reckoning capability, which guarantee a robust localization in places like a park where GPS has poor signal quality due to the obstruction from a dense grove. The GPS with good signal quality, in turn, can limit the drift of the dead-reckoning estimate error. The robot pose can be determined by fusing these sensors using an EKF (Extended Kalman Filter) algorithm [16]. The robot pose comprises two-dimensional planar coordinates relative to the world coordinate frame (i.e., a map), along with its angular orientation (since the robot is used on the grass, it is reasonable to consider that the robot operating in planar environments).

#### 2) Navigation module

The common navigation module includes path planning, path following and obstacle avoiding. Given an environment map, a path planner should generate a trajectory that can make the robot pass through the whole area with optimal performance cost, such as the shortest path or the minimum repeat coverage. The path planner proposed in [17] can generate a human-mimic path, which minimizes the path length and the sum of angle changes, and increases the battery running time. Thus, it is used for planning a path to cover the whole cleaning area. A path follower ensures the robot follow the planned path as closely as possible. Meanwhile, an obstacle avoidance module is needed to prevent the robot from colliding with any obstacles. We use similar strategy as proposed in [18] to address the path following and obstacle avoiding problems at the same time. It should be noticed that the obstacle avoiding here takes the floor segmentation results and the ultrasonic rangefinder measurement as input. The detailed obstacle avoiding algorithm is described in section III.C.

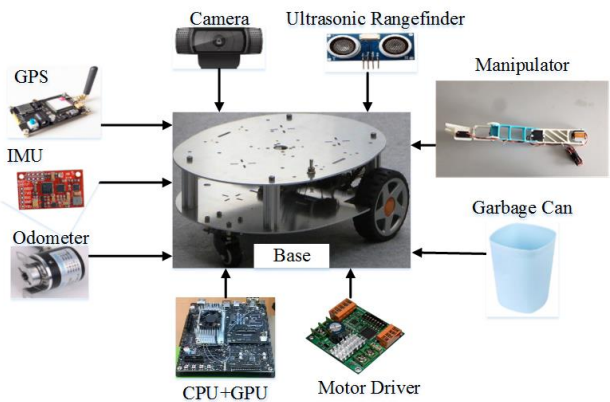

Fig. 2. The hardware configuration of the robot.

Besides, an object tracker is proposed for tracking the detected object. Once the perception module has detected objects in the image, the general navigation module will be suspended, and the object tracker will command the robot to approach the specific object gradually. When the robot is in a suitable position for picking up the object, the robot will cease to move and start to determine if the object is garbage. If yes, the manipulator will pick it up; otherwise, the obstacle-avoiding module will control the robot to bypass it. Then, the general navigation module will resume controlling the robot to follow the planned path for cleaning continually.

3) *Perception module*

The perception module includes ground segmentation, garbage recognition and localization. The ground segmentation is used to detect object and ground. The garbage recognition determines whether the detected object is garbage (see Section III.A). If the object is garbage, the garbage localization will provide the coordinates of the garbage in the image.

4) *Driver module*

Both the speed of the robot and the action of the manipulator are driven by this module. The motor driving technology is already off-the-shelf, and many low-cost motor drivers on the market can be used on this robot.

5) *Map module*

This module is only used in the early stage for constructing the environmental map, which guides the robot to perform the cleaning task. It does not consume any extra computation cost during operation. The map is represented by the occupied grid [19], which is a fine-grained grid over the continuous space of locations in the environment. The map construction only requires the boundary of the cleaning area, which can be obtained from the GPS data and projected on the map.

## III. Key Methodology

### A. Perception

With the rise of deep learning, the CNN improves the image comprehension by learning more discriminative and richer features. The proposed robot should be capable of recognizing

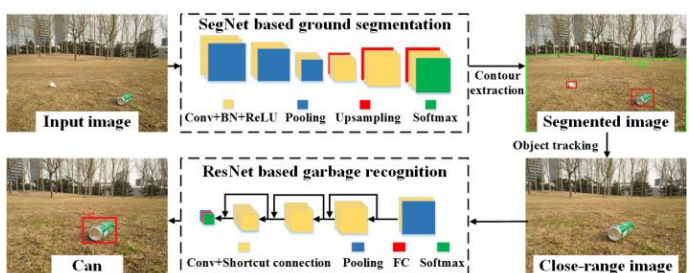
Fig. 4. Processing flow of perception.

the garbage and other obstacles (namely, determine the passable area) in the image. Through the pixel-wise semantic labels produced from SegNet [20], the ground can be distinguished from the other areas or objects. However, the object on the ground cannot be recognized exactly because the scene was just coarsely segmented. Thus, an image classification network is required. ResNet [21] can be a good option for object classification and object localization due to its residual learning framework. As shown in Fig. 4, an image was firstly inputted into the ground segmentation, which is based on SegNet. Then, the segmented image can be acquired. The segmented image (see Fig. 5) can provide the area of ground and the object on the ground. If there is an object on the ground (see Fig. 5(a)), the object tracker (see section III.B) will control the robot to approach the object. Besides, multiple objects maybe exist in the ground contour (see Fig. 5(b)). In this case, the closest one will be selected to track. Then, the close-range image can be acquired and used for object recognition. An instance of resulting image after recognition is shown in Fig. 6. If the object is recognized as garbage, the manipulator will pick it up; otherwise, the robot will take it as an obstacle and plan to avoid it.

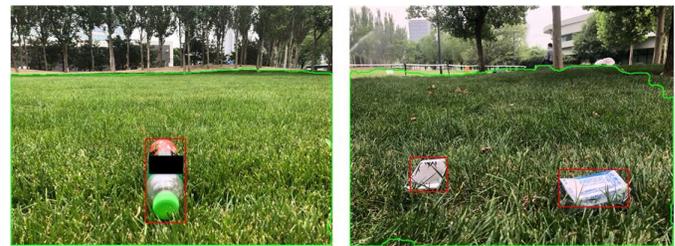
(a) Single object        (b) Multiple objects
Fig. 5. Example of the ground segmentation (the area of green line is the ground; the object in the red box is the potential garbage).

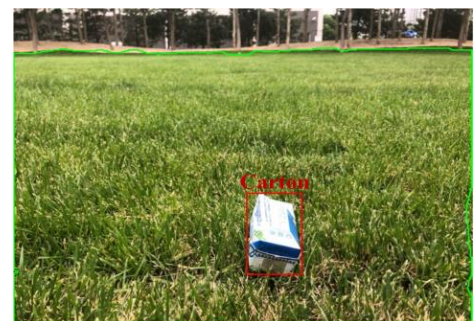
Fig. 6. Example of the garbage recognition.

### B. Object Tracker

Once the object on the grass was detected or segmented by the SegNet, the robot will be controlled to approach the object for recognizing it more clearly and picking it up appropriately by the proposed object tracker. The object tracking can be realized through the following steps:

Firstly, the boundary (the red rectangle in Fig. 5) coordinates of the object in the image can be obtained by the segmented image, and denoted by $\begin{bmatrix} u_{tl} & v_{tl} \\ u_{br} & v_{br} \end{bmatrix}$, where $(u_{tl}, v_{tl})$ represents the pixel coordinates of the top-left corner and $(u_{br}, v_{br})$ represents the pixel coordinates of the bottom-right corner. Thus, the coordinates of the object center are $(\frac{u_{tl}+u_{br}}{2}, \frac{v_{tl}+v_{br}}{2})$, expressed as $(u_{cO}, v_{cO})$. The bottom-center coordinates of the image can be represented by constant values $(u_{cI}, v_{cI})$ (If the image resolution is 640*480, $(u_{cI}, v_{cI}) = (319, 479)$).

Secondly, compute the difference between the object and the bottom-center of the image, which is given by

$$\Delta u = u_{cI} - u_{cO}$$
$$\Delta v = v_{cI} - v_{br}. \quad (1)$$

Thirdly, compute the speed of the robot according to $\Delta u, \Delta v$:

$$v = \begin{cases} \dfrac{\Delta v}{h} v_m & \text{if } \|\Delta v\| > 10 \\ 0 & \text{else} \end{cases}, \quad (2)$$

$$\omega = \begin{cases} \dfrac{\Delta u}{w/2} \omega_m & \text{if } \|\Delta u\| > 5 \\ 0 & \text{else} \end{cases}, \quad (3)$$

where $v, \omega$ is the translation and angular speed respectively; $v_m, \omega_m$ represents the maximum translation and angular speed respectively; $h, w$ is the height and the width in pixels of the image.

The above 3 steps are performed continuously, and when the object approximately locate in bottom-center of the image, the robot stop to recognize the object and then determine whether picking up the object or not.

*C. Navigation*

Most of the existing road sweeper trucks or vacuum cleaning robots follow a pre-defined path [17] to clean trashes without distinguishing between the garbage and other objects. When operating on the grass, as is shown in Fig. 7, these robots may clean the unnecessary trash (e.g. dust, leaves) or non-garbage (e.g. mobile phone, shoes). Thus, the robot must dynamically plan the path according to the navigation goal (i.e. the garbage to pick up), throughout the entire cleaning task. This paper, based on garbage recognition and ground segmentation, designs a novel navigation strategy (see Fig. 8), which is depicted in the following steps:

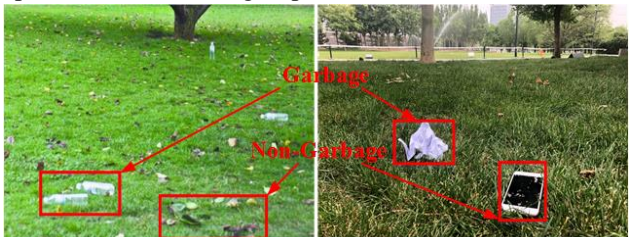

Fig. 7. Examples of working environments.

Step 1: according to the robot location and the map, it judges whether the robot is out of the map. Because the map is a binary occupied grid (the black pixel represents occupied space and white is free, see Fig. 9), and the location of the robot is mapped on the occupied grid in real time, it is easy to determine if the robot is out of the boundary (the pixel value 0 means the robot is out, and 1 is not).

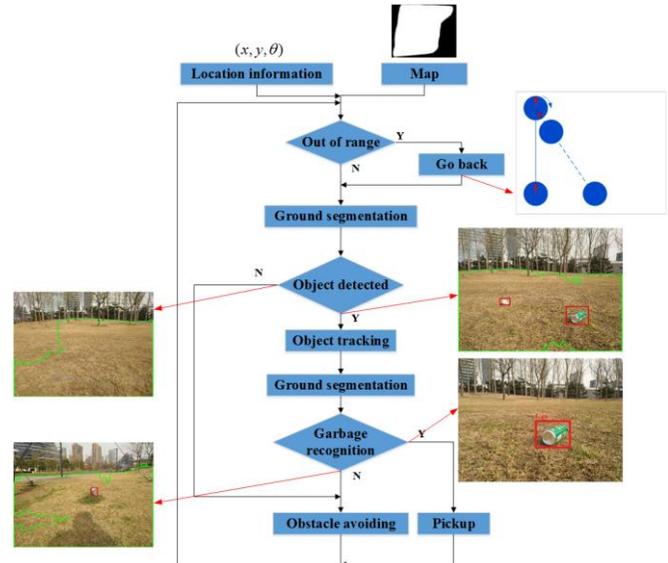

Fig. 8. Processing flow of navigation.

Step 2: if the robot is out of the map, then it will be controlled to go back; otherwise, go to Step 3. The return mechanism is that the robot rotates a random angle and then checks the pixel values (the number of the searching pixels depend on the resolution of the map) in the robot's heading direction after rotated. If all pixels are 1, the robot goes ahead; otherwise, the robot continues rotating. The above process can intuitively be depicted in Fig. 9. This is similar to the random walk strategy that is adopted to many existing vacuum cleaning robots.

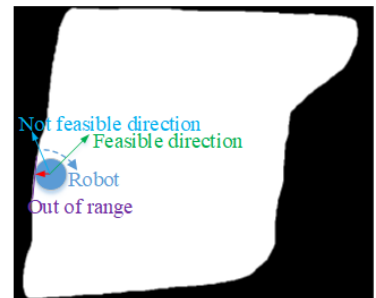

Fig. 9. The return mechanism of the robot.

Step 3: the ground segmentation results can provide the ground contour and the objects in this contour. If some objects were detected, then the object tracker (see section III.B) will work to make the robot approach this object (go to Step 4); otherwise, the robot will find the feasible direction according to the ground contour (go to Step 5).

Step 4: when the object approximately locates in bottom-center of the image, the object will be determined as garbage or common obstacle by the garbage recognition. If the object is regarded as garbage, the manipulator will work for picking the object up; otherwise, the obstacle avoiding will run to avoid this obstacle (go to Step 5).

Step 5: based on the ground segmentation results, the obstacle avoiding will produce the optimal direction. The robot will follow this direction to go on its cleaning task. The optimal

0000-0003-4322-6598　　　　　　　　　　　　　　　　　　　　　　　　　　　　　　　　　　　　　5

direction can be given by the following description (see Algorithm 1).

1.) No object in the ground contour. Firstly, according to the contour of the ground, compute the center pixel of the feasible area every row in the image. As is shown in Fig. 10, it is possible that many passable intervals exist in a row (see the blue line in Fig. 10). Thus, the widest interval will be selected and compute the center pixel value $(u_c, v_c)$. In practice, only half of the image pixels (from middle row (see the red line in Fig .10) to the last row) are required to provide the feasible direction. Many center pixel values will be produced, denoted by $Cent = \{(u_{c_i}, v_{c_i}) | i = 1, \cdots, n\}$, where n is half the numbers of the image rows.

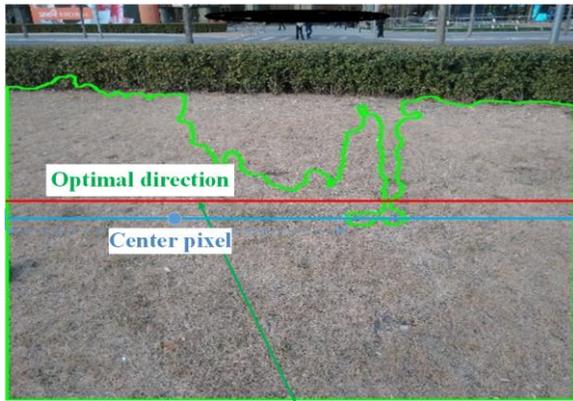

Fig. 10. Example of computing the optimal direction (without object).

2.) Object exists in the ground contour. The computing process is almost the same with 1.), the difference is that the object boundary should take into account. As shown in Fig. 11, the center pixel value is the result after the ground boundary subtracted the object boundary in a row (see the blue line in Fig. 11). Many center pixels also can be obtained, and represented as $Cent = \{(u_{c_i}, v_{c_i}) | i = 1, \cdots, n\}$, where n is half the numbers of the image rows.

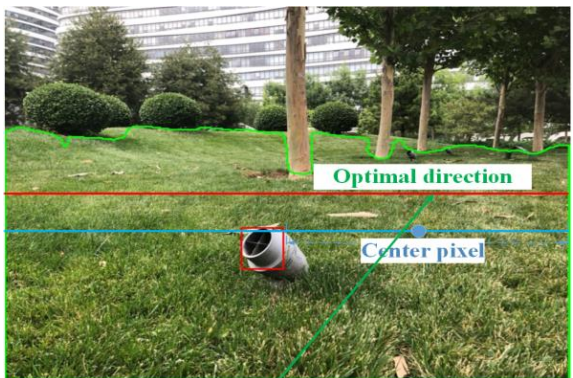

Fig. 11. Example of computing the optimal direction (with object).

Based on the center pixels $Cent = \{(u_{c_i}, v_{c_i}) | i = 1, \cdots, n\}$, the optimal direction can be fitted through Hough Transform [22] (function hough_line() in Algorithm 1). Therefore, the optimal direction (see the green arrow in Fig. 10 and Fig. 11) in the image can be obtained. And it can be easily transformed into the real motion direction of the robot according to homography mapping (homography matrix H in Algorithm 1) between the image and the planar ground.

**Algorithm 1** Find Optimal Direction

initialize vector *In, Cent*

for i=0, h/2-1 do

　initialize in_w =0, max_w=0

　for j=0,w-1 do

　　determine if point *P* (j, i-1+h/2) is in the ground contour *GC*

　　flag_gc= pointPolygonTest[1] (*GC*, *P*, false)

　　if flag_gc=1, *P* is inside *GC*, then

　　　determine if point *P* is in all the obstacles contours *OC*[M]

　　　for n=1,M do

　　　　flag_oc= pointPolygonTest[1] (*OC*[n], *P*, false)

　　　　if flag_oc=1, *P* is inside *OC*[n], then

　　　　　if *In*≠Ø, the interval width in_w = j-1-*In*[0]

　　　　　　if in_w> max_w then

　　　　　　　max_w=in_w, store ((j-1+*In*[0])/2, i-1+h/2) as *Cent*[i], clear *In*

　　　　　　else clear *In*

　　　　　break

　　　　if n==M, *P* is outside all the contours *OC*, then store j in vector *In*

　　end for

　　if flag_gc<=0, *P* is outside *GC*, then

　　　if *In*≠Ø, the interval width in_w = j-1-*In*[0]

　　　　if in_w> max_w then

　　　　　max_w=in_w, store ((j-1+*In*[0])/2, i-1+h/2) as *Cent*[i], clear *In*

　　　　else clear *In*

　end for

end for

[ρ,θ]= hough_line(*Cent*)

random select two points *A, B* on the line

maping point *A, B* to *A′, B′* on the real ground plane, $A'=H^{-1}A$, $B'=H^{-1}B$

$$optimal\ direction\ \theta' = \arctan(\frac{y_{B'} - y_{A'}}{x_{B'} - x_{A'}})$$

[*1] https://opencv.org/

The above five steps are operated repeatedly for cleaning the garbage in the whole area.

IV. EXPERIMENTAL RESULTS AND DISCUSSIONS

A. *Recognition Accuracy*

The key to pick up the garbage is the garbage recognition. If the recognition module fails to recognize the garbage, or recognize a non-garbage object as the garbage, the robot will work improperly. Thus, the recognition accuracy was firstly tested. The dataset consists of 40k training images and 7k testing images in six classes (5 garbage class and 1 non-garbage class). We present experiments trained on the training set and evaluated on the test set. The architectures of the network was used ResNet-34 [21]. We start with a learning rate of 0.01, divide it by 5 at 20k, 40k and 60k iterations, and terminate training at 75k iterations. The final testing results are shown in TABLE I.

TABLE I
CLASSIFICATION ERROR ON TEST SET

| Category | Error[*1] (%) |
|---|---|
| Bottle | 8.13 |
| Can | 9.89 |
| Carton | 9.06 |
| Plastic bag | 14.32 |
| Waste paper | 22.3 |

[*1] Predict one category of an image, it is not the same with the manual annotation category.

The error of the waste paper is obviously higher than the error of other category due to the lack of visual features. Similarly, some transparent plastic bags make the error of the plastic bag relatively high because transparent objects have less features and the deep neural networks have difficulty in recognizing these objects. However, in practical test, the success rate of picking the garbage up can be up to 96%. As long as the robot recognizes the object as the one of the above five categories, it will pick the object up. Therefore, even though the robot mistakenly considers the object as another one, the success rate of picking the garbage up is higher than the recognition accuracy. This demonstrated that the proposed robot could be used for autonomously cleaning the garbage in the real world.

In order to test the ability to discriminate between garbage and non-garbage (e.g., a book, a wallet, a phone, et al.), we use 6 common objects (cup, book, shoes, phone, bag, wallet) which may fall on the grass, and collect 750 images for testing. The result is shown in TABLE II. Although some objects have comparatively higher recognition error, we can set a larger probability threshold to avoid picking it up when the robot operates in real environment.

TABLE II
CLASSIFICATION ERROR ON NON-GARBAGE

| Category | Bottle | Can | Carton | Plastic bag | Waste paper |
|---|---|---|---|---|---|
| Cup | 0.153 | 0.184 | 0.012 | 0.009 | 0.003 |
| Book | 0.002 | 0.010 | 0.136 | 0.005 | 0.012 |
| Shoes | 0.005 | 0.023 | 0.038 | 0.009 | 0.003 |
| Phone | 0.007 | 0.011 | 0.065 | 0.004 | 0.008 |
| Bag | 0.007 | 0.013 | 0.009 | 0.032 | 0.004 |
| Wallet | 0.010 | 0.023 | 0.089 | 0.012 | 0.009 |

### B. Computational Cost

The average computational time for the key algorithm is calculated, and the results are shown in TABLE III. The ground segmentation and garbage recognition operate on GPU and takes about 10.3 ms and 8.1 ms respectively. Although this can be operated in real time, the garbage recognition, in practice, was not always operated. Only when the robot approaches the object, will the garbage recognition run. The obstacle avoiding runs on CPU and takes about 16.5 ms. The test results verify that the proposed algorithm can satisfy the real-time requirement. Because the garbage pickup is executed when the robot stopped, the real time requirement is not very high. The pickup costs about 1.4 s, which is relatively efficient for cleaning task. However, the time cost of the pickup is expected to be smaller.

TABLE III
COMPUTATIONAL TIME FOR THE KEY ALGORITHM

| Processing Step | Average Time |
|---|---|
| Ground segmentation[*] | 10.3 ms |
| Garbage recognition[*] | 8.1 ms |
| Obstacle avoidance | 16.5 ms |
| Pickup | 1.4 s |

[*] run on GPU

### C. Cleaning Efficiency Test

To test the cleaning efficiency, we have conducted two experiments on a playground (85.4 m * 73 m). Experiment 1 is that the robot cleans the whole area according to the planning path, as shown in Fig. 12; Experiment 2 is that the robot randomly travels to clean the whole area, as shown in Fig. 13.

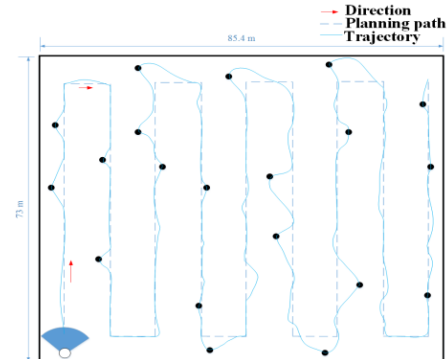

Fig. 12. Experiment 1: clean the whole area with planning path (with 20 garbage).

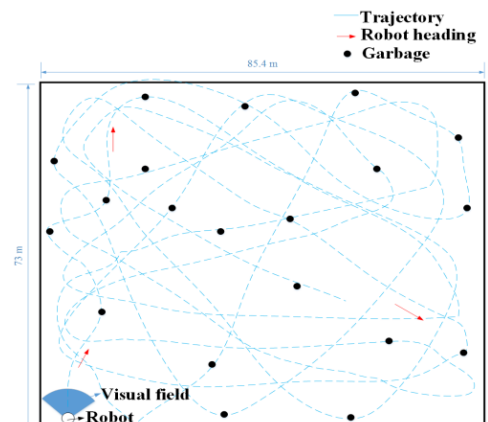

Fig. 13. Experiment 2: clean the whole area randomly (with 20 garbage).

The robot can perceive the range of 60° with about 10 m. We drop 20 and 50 garbage randomly on the playground for test. At different periods, the number of picked garbage was counted, as shown in TABLE IV. Experiment 1 takes about 42 minutes to pick all 20 garbage up; while Experiment 2 takes about 58 minutes. Because the path planning can cover the whole area optimally from the perspective of time cost, Experiment 1 can clean the whole area more efficiently than Experiment 2. However, if the numbers of the garbage increases, as shown in TABLE IV, Experiment 1 takes about 61 minutes and Experiment 2 takes about 65 minutes. The efficiency of Experiment 2 is approximately the same with Experiment 1. However, Experiment 2 is easier to implement than Experiment 1 because Experiment 2 has no need of path planning to coverage the whole cleaning area.

TABLE IV
NUMBERS OF THE GARBAGE AT DIFFERENT PERIODS

| Periods | | Numbers of garbage | |
|---|---|---|---|
| | | 20 in total | 50 in total |
| 10 minutes | Experiment 1 | 4 | 9 |
| | Experiment 2 | 4 | 8 |
| 20 minutes | Experiment 1 | 10 | 15 |
| | Experiment 2 | 7 | 13 |
| 30 minutes | Experiment 1 | 14 | 27 |
| | Experiment 2 | 9 | 28 |
| 40 minutes | Experiment 1 | 19 | 36 |
| | Experiment 2 | 12 | 37 |
| 50 minutes | Experiment 1 | 20[*1] | 45 |
| | Experiment 2 | 14 | 43 |
| 60 minutes | Experiment 1 | 20 | 49 |
| | Experiment 2 | 20[*2] | 48 |
| 65 minutes | Experiment 1 | - | 50[*3] |
| | Experiment 2 | - | 50 |

[*1] end on 42 minutes
[*2] end on 58 minutes
[*3] end on 61 minutes

## V. CONCLUSION

This paper presents a novel robot system for cleaning the garbage on the grass automatically. Based on the powerful deep neural networks, the proposed robot can recognize and pick up the garbage without any human assistance. Besides, the novel navigation algorithm based on the ground segmentation was proposed. Through the manipulator, relatively large garbage on the grass can be picked up. This cleaning mechanism is more suitable for cleaning the garbage on the grass than the one used by the existing road sweeper truck or vacuum cleaning robot. Experimental results proved that the proposed robot could recognize the garbage accurately and move efficiently. This robot can serve as a powerful tool for cleaning the garbage on a big lawn in a park or school.

[22] Z. Xu, B. S. Shin and R. Klette, "Accurate and robust line segment extraction using minimum entropy with Hough Transform," *IEEE Trans. Image Process.*, vol. 24, no. 3, pp. 813-822, Mar. 2015.

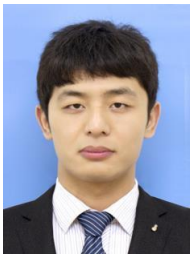

**Jinqiang Bai** received the B.E. degree and M.S. degree from China University of Petroleum (East China), Shandong, China, in 2012 and 2015, respectively. He is currently pursuing the Ph.D. degree with Beihang University, Beijing.

His current research interests include computer vision, deep learning, robotics and AI.

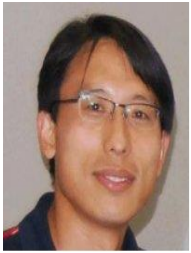

**Shiguo Lian** (M'04) received the Ph.D. degree from Nanjing University of Science and Technology, China.

He was a research assistant in City University of Hong Kong, Hong Kong, in 2004. From 2005 to 2010, he was a Research Scientist with France Telecom Research and Development Beijing, Beijing, China. He was a Senior Research Scientist and Technical Director with Huawei Central Research Institute, Beijing, China, from 2010 to 2016. Since 2016, he has been a Senior Director with CloudMinds Technologies Inc., Beijing, China. He has author over 80 refereed international journal papers covering topics of artificial intelligence, multimedia communication, and human computer interface. He has authored and co-edited over 10 books, and held over 50 patents.

Dr. Lian is on the editor board of several refereed international journals.

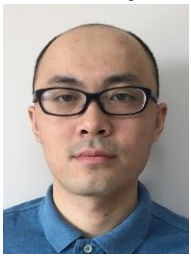

**Zhaoxiang Liu** received the B.S. degree and Ph.D. degree from the College of Information and Electrical Engineering, China Agricultural University, Beijing, China, in 2006 and 2011, respectively.

He joined VIA Technologies, Inc., Beijing, China, in 2011. From 2012 to 2016, he was a Senior Researcher with the Central Research Institute of Huawei Technologies, Beijing, China. He has been a Senior Engineer with CloudMinds Technologies Inc., Beijing, China, since 2016. His current research interests include computer vision, deep learning, robotics, and human computer interaction.

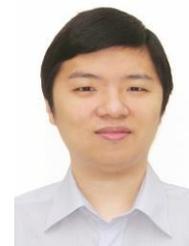

**Kai Wang** received his Ph.D. degree from Nanyang Technological University, Singapore in 2013.

He was a Senior Researcher with the Central Research Institute of Huawei Technologies, Beijing, China. He has been a Senior Engineer with CloudMinds Technologies Inc., Beijing, China, since 2016. He has published over ten papers on international journals and conferences. His research interests include augmented reality, computer graphics, and human-computer interaction.

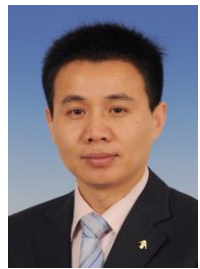

**Dijun Liu** received the Ph.D. degree from China University of Petroleum (East China), Shandong, China.

He has been the Chief Scientist of China Academy of Telecommunication Technology, Beijing, China, and a Professor of Beihang University, Beijing, China. He was the Director of China Institute of Communications. He has over 20 years of prized academic research, industrial development, and entrepreneurship (Chief Scientist, Vice President, and CEO) in semiconductor and communication.

Dr. Liu was a recipient of the 2016 National Science and Technology Progress Special Award by China State Council. He was the Chairman of China Communications Integrated Circuit Committee.